\documentclass{article}
\usepackage{amsmath,amsfonts,amssymb,times,graphicx,subfigure,natbib,algorithm,algorithmic,hyperref}
\usepackage{times}
\pdfoutput=1

\usepackage[accepted]{data4good2016}

\icmltitlerunning{Designing Intelligent Automation based Solutions for Complex Social Problems}
\begin{document} 

\twocolumn[
\icmltitle{Designing Intelligent Automation based Solutions for Complex Social Problems}

\icmlauthor{Sanjay Podder}{sanjay.podder@accenture.com}
\icmladdress{Accenture Labs, Bangalore 560066 India}
\icmlauthor{Janardan Misra}{janardan.misra@accenture.com}
\icmladdress{Accenture Labs, Bangalore 560066 India}
\icmlauthor{Senthil Kumaresan}{senthil.kumaresan@accenture.com}
\icmladdress{Accenture Labs, Bangalore 560066 India}
\icmlauthor{Neville Dubash}{neville.dubash@accenture.com}
\icmladdress{Accenture Labs, Bangalore 560066 India}
\icmlauthor{Indrani Bhattacharya}{indrani@cinindia.org}
\icmladdress{CINI, Kolkata, India}


\vskip 0.3in
]

\begin{abstract} 
Deciding effective and timely preventive measures against complex social problems affecting relatively low income geographies is a difficult challenge. There is a strong need to adopt intelligent automation based solutions with low cost imprints to tackle these problems at larger scales. Starting with the hypothesis that analytical modelling and analysis of social phenomena with high accuracy is in general inherently hard, in this paper we propose design framework to enable data-driven machine learning based adaptive solution approach towards enabling more effective preventive measures. We use survey data collected from a socio-economically backward region of India about adolescent girls to illustrate the design approach.   
\end{abstract} 

\section{Introduction}
\label{intro}
Relatively low income societies in countries like India face multitude of challenges~\cite{Ref6} including low empowerment of weaker sections of society, poor health and low nutrition, low quality of education, poor child protection, and poor quality of sanitation and hygiene.

To address these challenges and associated problems like child trafficking~\cite{sarkar2014rethinking,aljazeera}, child labor, domestic violence etc. community outreach program (COP) has emerged as one of the prominent methods adopted in many of the Low and Middle Income Countries (LMIC) to deliver healthcare services, monitor the condition of a vulnerable population, and propagate information during disasters and so on. With the advent of Social, Mobile, Analytics and Cloud based digital technologies, and the penetration of smartphones in rural areas have enabled many organizations to upskill the community facilitators (CFs) through digital technologies. Existing literature~\cite{Ref3,Ref1} has highlighted the challenges in implementing a digital solution for delivering high-quality outreach services and the system architecture guidelines that have to be followed to overcome those challenges. Since, the internet penetration in rural areas is still relatively low, organizations have started to adopt a Mobile based Decision Support System (MDSS) that can work without any internet connectivity. MDSS have helped many organizations catering to outreach care, with in-built rule sets to categorize the target population and ease the work of outreach workers from complex analysis based upon multiple guidelines.  

We argue in this paper that instead of a fixed rule set, how a machine-learning based, dynamic and context aware computational model could help to provide improved quality of care through the community outreach programs. However, the primary challenge in applying such data-driven approaches in the context of social problems is the lack of verifiable and quality data. Government and researchers rely on NGOs’ that are delivering outreach services as a primary sources of data. But, primary data collected by many organizations may be unreliable and most of the information is collected to meet only organizational needs and may be unsuitable otherwise. 

\subsection{Example Scenario: Child Trafficking} 30 Million people are trafficked globally every year. Traffickers often utilize mobile phones, social media, online classifieds, and other networking channels to interact with their circles about the victims. To counter this, many NGOs like CINI (Child In Need Institute) India~\cite{Ref7} are leveraging field agents to monitor the vulnerable population of adolescent girls between the age group of 10 and 19. CINI field agents periodically collect data about vulnerable girls on aspects pertaining to verticals like education, protection, health, and nutrition and analyze the data along all these vertical together to determine potential vulnerabilities. 

\section{Proposal for a Data Driven Dynamically Adaptive Design Framework} 
To solve complex social problems~\cite{Ref3,Ref5} particularly affecting relatively low income geographies, there is greater need to adopt ‘intelligent computing solutions’ involving minimal cost imprint with maximum empowerment of the potential victims (PVs) which remain under powered owing to various socio-economic factors. We will refer such AI driven platforms or applications designed for Social Good to address complex social challenges as AI4SG.

Next we argue for adoption of certain design themes while developing these AI4SG applications and platforms. For illustration, we use survey data collected by CINI from a socio-economically backward region of India about 1000 adolescent girls towards their vulnerability analysis and consequent mitigation. Details of this can be found in earlier published study~\cite{Ref2}. Illustrative analysis of the data was carried out using R 3.2.4.

\subsection{Design Thinking Proposals}
Social phenomena are inherently hard to model accurately~\cite{Ref3,Ref5}. Primary reason for this could be attributed to large number and variety of factors affecting the phenomena under study in ways too complex to be fully understood. To further complicate the matter in the context of social problems, for ethical reasons, controlled studies cannot be performed since actual social events cannot be artificially created but could only be analyzed when they occur naturally. Therefore approaches applying static and analytical solutions (e.g., closed form formula based vulnerability analysis~\cite{Ref2}) cannot reliably generalize to larger contexts and might remain locally relevant where most of the parameters in the model are approximately fixed and attributes with high predictive power are known with experience. 
\begin{figure}[ht]
\begin{center}
\centerline{\includegraphics[width=\columnwidth]{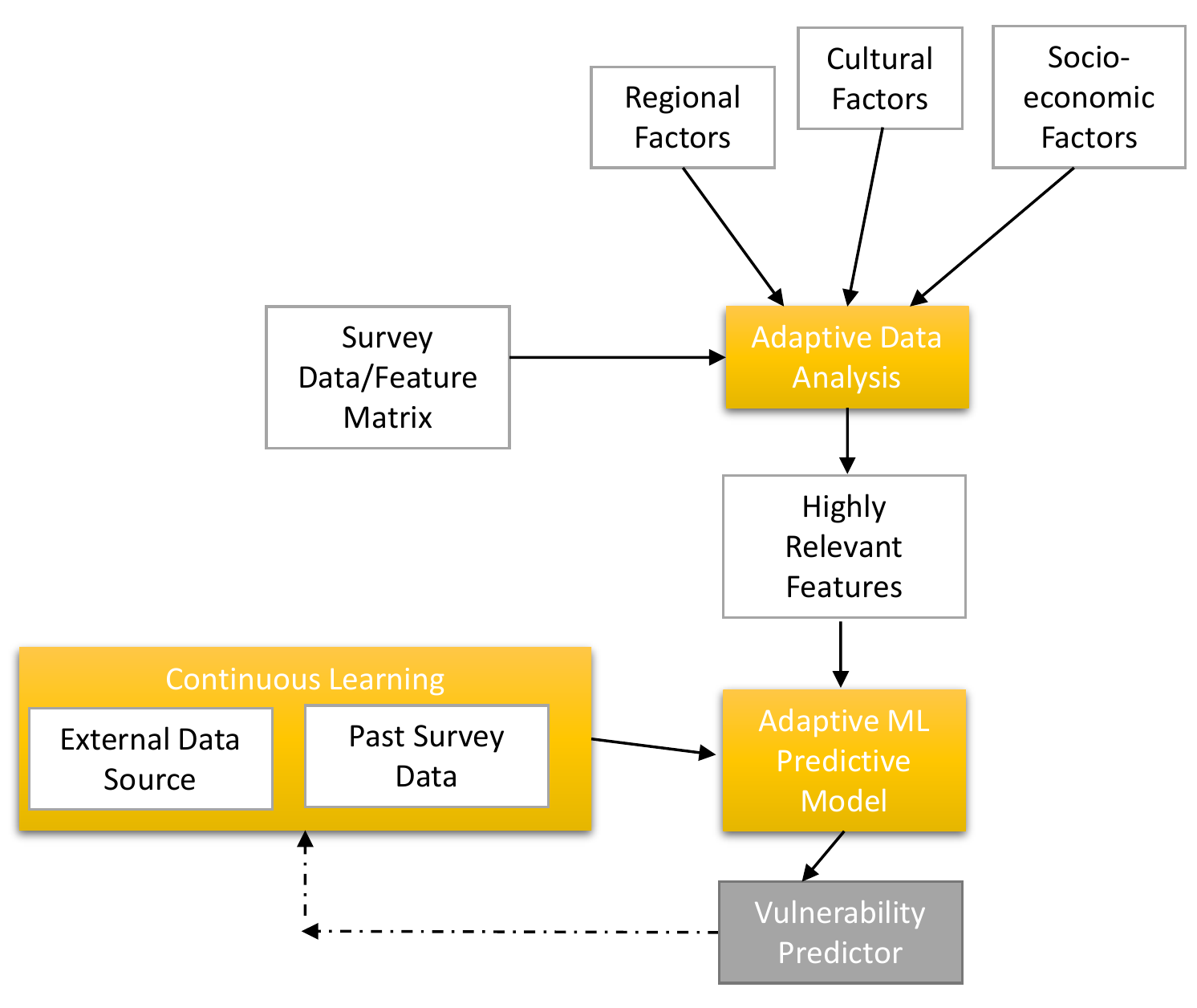}}
\caption{High Level Component Architecture of the Proposed Design Approach for the problem of Child Trafficking}
\label{Fig1}
\end{center}
\vskip -0.2in
\end{figure} 
When generalization beyond local boundaries and large scale adoption are critical goals to achieve, a data-driven and machine learning based approach may provide an effective resolution to this problem. Under such design framework, a computational model is generated (instead of a manually defined analytical model) from sample data collected from the field studies with feature-set designed in consultation with social scientists specializing in that field.

We aim to evolve an approach to build a framework to design such models for various social problems. Framework is primarily based upon data-driven design methodology together with application of AI technologies to render eventual solution amenable to wider adoption with low cost imprint and serving priorities at multiple levels ranging from potential victims (e.g., children as potential targets of trafficking) to field workers, to NGOs and Government Agencies interested in analysis of impact of their services, and eventually to social scientist interesting in scientifically studying the underlying phenomena at larger scales. 

For notational convenience, we will use community facilitator, field agent, and agent interchangeably. 

\subsection*{Design Proposal 1: Eventual data-driven machine learning based predictive modelling}
Analytically designing solutions for complex social problems is inherently hard and only effective alternative is to design a model which optimally conforms to the data collected from the field. Machine learning provides an operational solution to this problem where patterns underlying the data which could provide clues to solving the problem can be computationally extracted and used in designing mitigation strategies. 

Often solving social problems requires an ability to make predictions well ahead of time before actual event may take place (e.g., vulnerability prediction for child trafficking problem) using analysis of factors affecting potential victims. In this perspective classification and regression techniques provide required predictive model though initial design trials may be necessary to determine the right prediction technique or a combination of many. 

However, acquiring sufficient good quality data to train machine learning models in the context of social problems is very difficult. For example, even though the collected survey data by CINI covers many interesting details~\cite{Ref2} pertaining to eventual vulnerability of children, it does not yet contain information about those who were actually reported to be trafficked. This may result into {\it cold-start} problem if only ML based model has to be used to design AI4SG applications. For this reason, this design proposal suggests that ML based data-driven solution should be the \textit{eventual} design goal and in order to start the application in field work by CFs (and PVs) and to gain their trust, one needs to have alternative heuristic solutions resulting from prior field experiences designed in collaboration with social experts. For example, for the problem of child trafficking~\cite{Ref2} discusses a linear convex model with 32 features along with a threshold to determine whether a child is vulnerable or not. This solution is being currently used by CFs of CINI to collect survey data and perform analysis.

However, it is important to add that with this proposal, we strongly suggest that except feature engineering, role of heuristic solutions should be minimized overtime as more data gets acquired on a real-time basis (see next proposal) so that final predictive model evolves by learning only from actual data and less from subjective experiences of solution designers to reduce biases and dependencies on AI4SG designers.    

\subsection*{Design Proposal 2: Real-time continuous learning based dynamic evolution of predictive model}
To motivate this design choice, let us consider a hypothetical scenario related to human trafficking use case. In this scenario, there has recently been cases of child trafficking in a locality during election time, however not all of those could be correctly predicted to be vulnerable by existing model. Therefore to update underlying prediction model, new data needs to be sent to its designers, which would then involve new cycle of update and reloading of the predictive model to the agent devices on periodic basis. Often such solutions even if built using ML techniques require centralized offline update of the predictive model and AI4SG applications running on agent devices cannot adapt themselves at run-time when new cases of actual victims become known! 

Towards that we suggest that solutions for social problems must be designed as continuously adaptive applications which learn (from potentially incomplete data) while being in actual use by retraining themselves automatically when information about new actual incidents is entered on the agent device running the application. Eventually overtime each agent would have evolved its own unique predictive model based upon the incidents of the trafficking known in her area and other cases where such trafficking did not take place for known period of time. Applications should also update their prior predictions after improved training and send alerts about all those, who now are now in danger zone but earlier were not. 

Additionally, agent device or central server should be designed to analyze updated field-data to infer which factors are becoming increasingly critical in the light of new incidents so that right mitigation strategies can be designed or existing ones could be adapted to meet the requirements of the emerging scenarios. For example, based upon these updated predictions, AI4SG application for child trafficking should send alerts to all the registered children (and/or their care takers) and community facilitators regarding changes in the mitigation strategies. 

\subsection*{Design Proposal 3: Structural analysis of data using feature correlations, similarities, and clustering}
Collecting details about actual victims of social problems is a known challenge~\cite{Ref3} – primarily because these victims are often out of access for any examination and only indirect data points could be collected with enough efforts. On the other hand, data for non-victims is relatively easier to acquire but it only makes design of prediction model harder owing to inherent bias towards non-victim class. Additional difficulty arises because when a prediction model used in practice, its predictions control mitigation strategies which further bias population towards its predictions and hence make it harder to know to what extent such a model is inherently accurate. 

Under such a scenario, unsupervised ML techniques should be used for complementary analysis of the data even as new data points get added. Examples of such analysis are considered next.

{\small \sf Cluster and Similarity Analysis:} Similarities among potential victims can be used to cluster them in social-groups by applying clustering techniques and to identify outliers. For example, a safety profile containing only those factors which may render a potential victim highly vulnerable could be defined and all the known PVs having similar profiles within same locality can be made to socially connect with each other so that they can work as a group  to address  their vulnerabilities together (see Figure~\ref{Fig6}). Similarly, clustering analysis can be used to determine whether certain details about a new PV are far away from others in the same locality? Note that in low income geographies, high levels of social similarities within same locality are a commonly observed phenomena. If so, AI4SG application alerts the agent with factors where high deviations are present. 

The similarity graphs or clusters can be further augmented with contextual knowledge about external environmental factors affecting the underlying phenomena (e.g., large scale religious gathering making trafficking of children easier for anti-social elements~\cite{aljazeera}). Such augmented graphs (type of knowledge graphs) can further assist in taking timely preventive measures as per the emerging contexts. 

For illustration, when we applied Hierarchical Clustering on CINI data in 17 dimensional PCA space, an outlier cluster with 7 sample points emerged which on closer analysis turned out to be involving invalid value ranges for certain attributes. Figure 2 illustrates the hierarchical clustering as dendrogram. 
\begin{figure}[ht]
\begin{center}
\centerline{\includegraphics[width=\columnwidth]{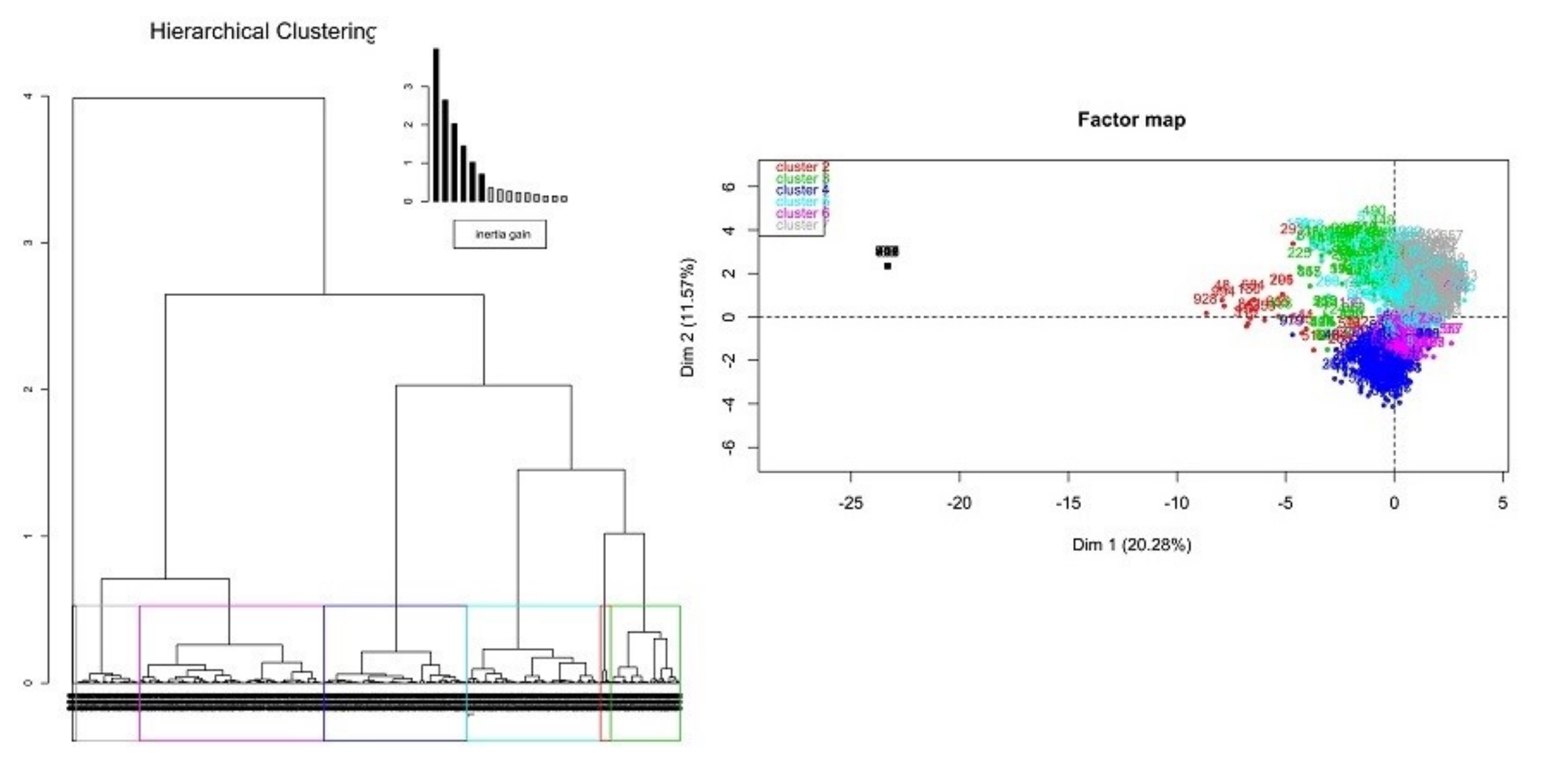}}
\caption{Hierarchical Clustering of Data Points with PCA Factor Map with 7 Clusters}
\label{Fig2}
\end{center}
\vskip -0.2in
\end{figure} 
Further {\small \sf similarity analysis} of the data revealed interesting insights on the nature of collected data samples. Figure~\ref{Fig4} illustrates this. For example, for approximately 48\% of the cases, each girl data had at least one another identical data point also present in the samples. Furthermore, the collected data turned out to be highly biased towards high similarity regions and lacking samples in other spectrum of similarities - there existed less than 5\% of pairs of girls with similarities less than 70\%. Thus indicating clear need for collecting survey data from multiple sources so that there are true representative samples present for designing predictive models and further analysis. 
\begin{figure}[ht]
\begin{center}
\centerline{\includegraphics[width=\columnwidth]{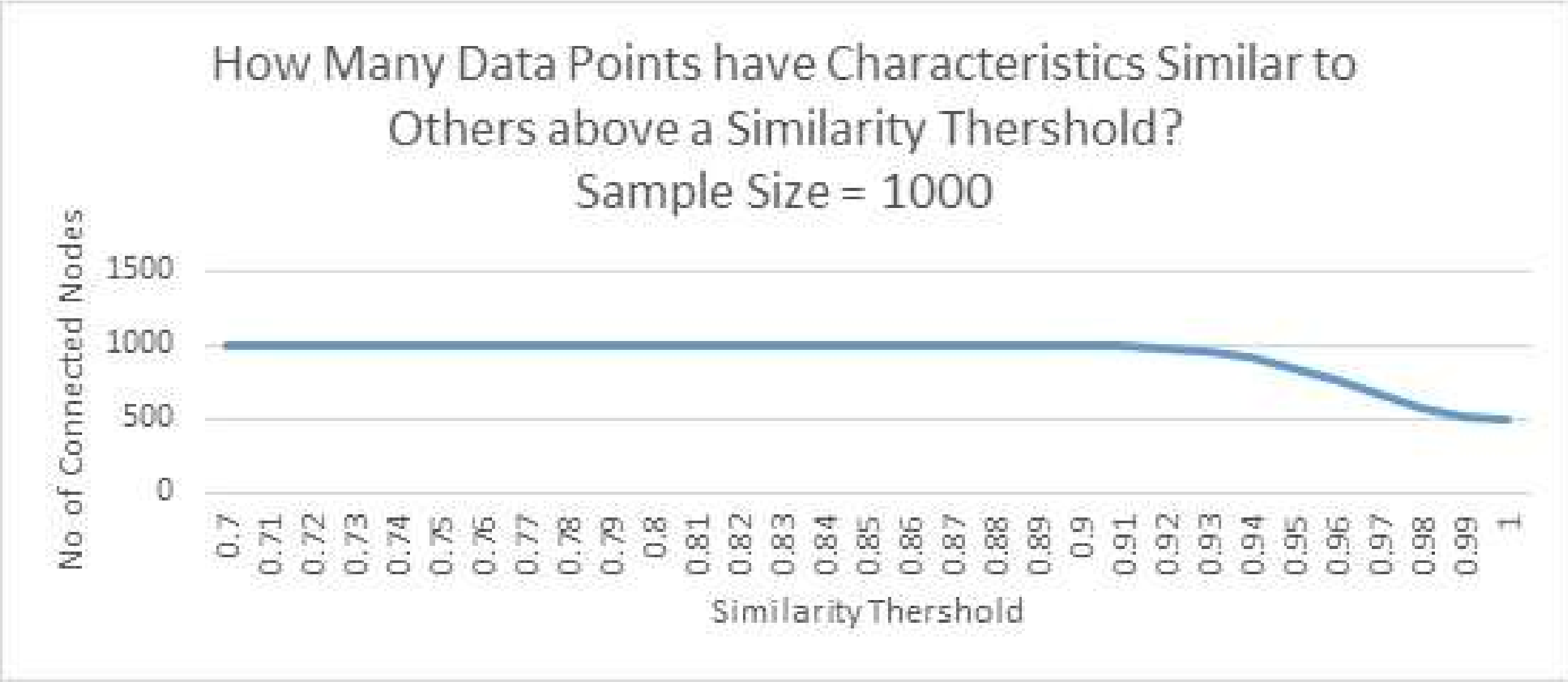}}
\caption{Identifying distribution of Data Points on Similarity Scale}
\label{Fig3}
\end{center}
\vskip -0.2in
\end{figure} 

{\sf \small Feature Correlation Analysis}: Comparison of positive and negative correlations among features can bring subtle insights into the intentional design of underlying choices. For example, Figure~\ref{Fig4} depicts the overall Correlogram with pair-wise correlations among the features (i.e., questions in the CINI survey data). In this correlogram features are ordered by first principal component loadings. To complement this Correlogram, Figure~\ref{Fig4} also depicts positive correlation graph among features as present in the survey data with correlation strength at least 0.5. 
\begin{figure}[ht]
\begin{center}
\centerline{\includegraphics[width=\columnwidth]{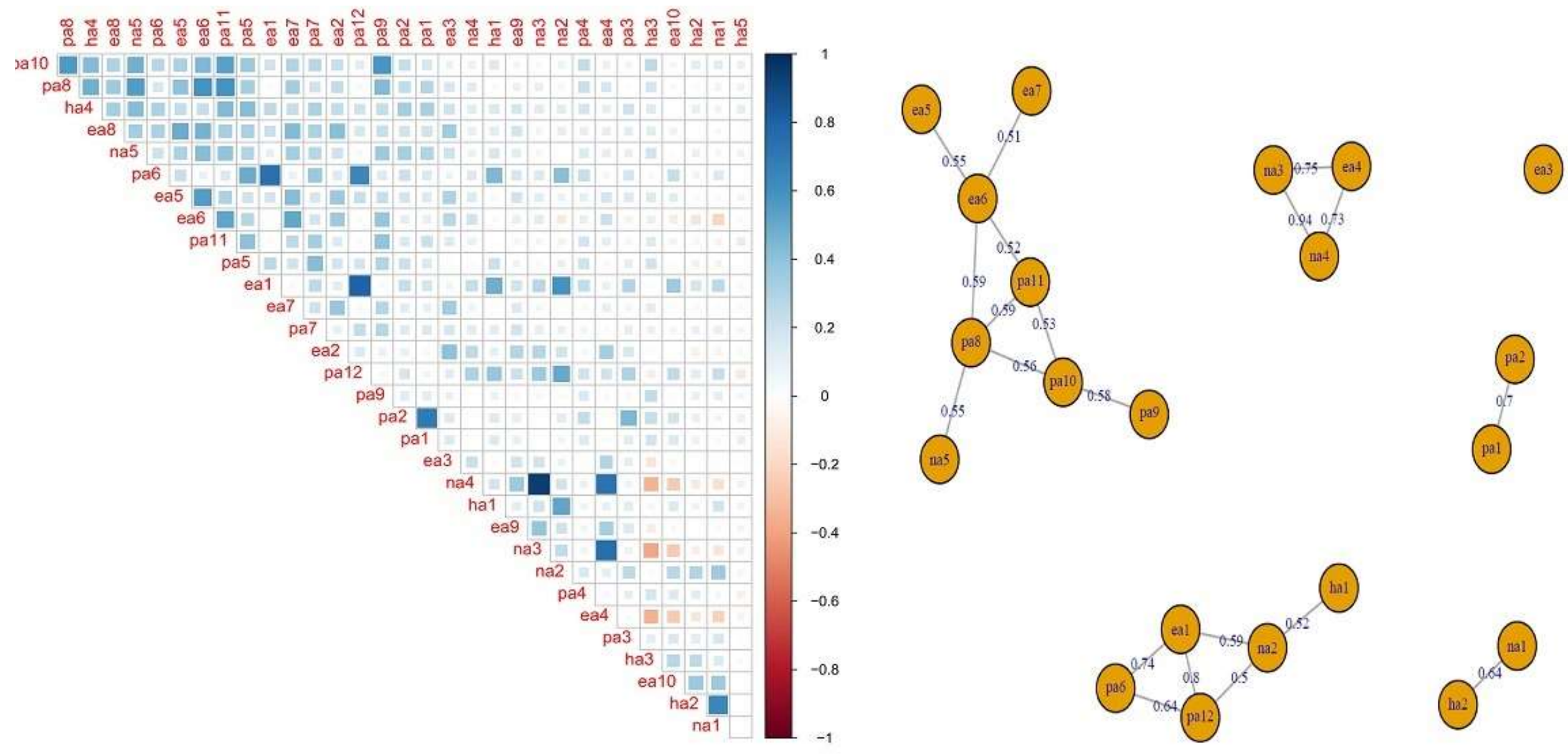}}
\caption{Correlogram depicting degree of associations among features and associated Positive Correlation Graph}
\label{Fig4}
\end{center}
\vskip -0.2in
\end{figure} 
This correlation analysis makes it clear as to which characteristics used in the underlying predictive model are actually related with each other and therefore possibly should be analyzed together. This correlation analysis should be further strengthened by Principal Component Analysis (PCA) for identifying most explanatory features. For example, from the CINI data, it turned out that 1st principal component had only 21\% explanatory power and hence any linear combination of features can only achieve at most this much explanatory capacity. Another interesting insight has been that it would require at least 17 PCs to achieve 85\% information variance in data implying that most of the originally designed survey questions for CINI data were largely uncorrelated with one another. This may shed further insights on the social dynamics of the underlying population being surveyed.

\subsection*{Design Proposal 4: Virtual agent based native spoken language interaction for decentralized empowerment of potential victims and for real-time data collection by CFs}
Generally community facilitators of COPs are required to in-person visit and collect data periodically about potentially vulnerable sections of society (e.g., potentially vulnerable children). This could be a critical impediment to scale such solutions to larger scale. 

This design proposal envisions the use of native written and/or spoken language based virtual agents executing on the mobile devices of these CFs as well as (if feasible) potential victims (or places like schools kiosks) so that CFs can interact with their MDSSs with relative ease and each registered user (e.g., girl child) having access to this virtual agent can send her current state to their assigned CF without requiring in-person interaction. Effectively, each virtual agent on the potential victim's device acts as a proxy to actual community facilitator and thus enables these CFs to get connected to larger number of potentially vulnerable victims (e.g., children) simultaneously.

Figure~\ref{Fig6} below depicts a virtual agent based hypothetical advisory scenario on enrollment of a new member by a community facilitator. Such virtual agents may also be used to authenticate users, reduce deliberate or unintentional fudging of data, and may enable emergency responses.
\begin{figure}[ht]
\begin{center}
\centerline{\includegraphics[width=\columnwidth]{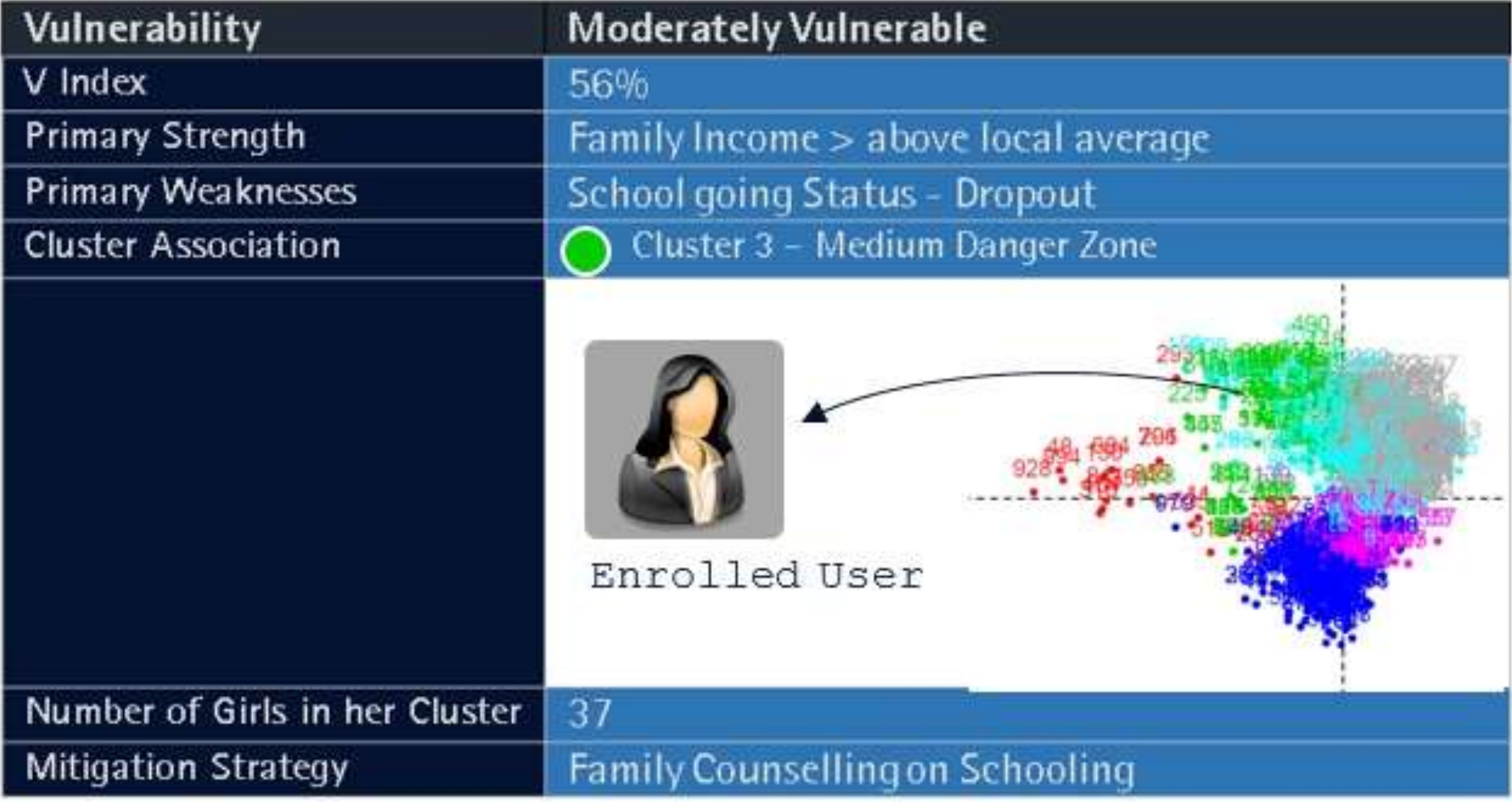}}
\caption{Illustrative Scenario where Virtual Assistant displays Vulnerability Analysis, Similarity based Clustering and recommends mitigation strategy to Community Facilitator on her mobile device}
\label{Fig6}
\end{center}
\vskip -0.2in
\end{figure} 
\section{Conclusion} 
To address the grand challenges of complex social problems in low income geographies, this paper presents design proposals to implement purely data-driven machine learning based solutions for enabling dynamic decision making towards deciding timely preventive measures which can be applied directly by field agents of community outreach programs. We primarily focused on design of real-time continuous learning based predictive application scenarios and structural analysis of data to enable fine grained analysis of local population a field agent is responsible for. This needs to be augmented with additional design elements including enabling large scale data processing for wide scale adoption~\cite{Ref4}, collective collaboration, and techniques for knowledge graph generation and their use in deciding preventive measures.      

\bibliography{rfs}
\bibliographystyle{icml2016}

\end{document}